\documentclass[10pt, a4paper]{article}

\usepackage[final]{lrec2026} 

\usepackage{longtable}
\usepackage{enumitem}
\usepackage{array}

\usepackage{booktabs}
\usepackage{fontawesome}
\usepackage{xcolor}

\title{Goldfish: Monolingual Language Models for 350 Languages}

\name{Tyler A. Chang$^1$, Catherine Arnett$^2$, Zhuowen Tu$^{1,3}$, Benjamin K. Bergen$^1$} 

\address{$^1$Department of Cognitive Science, $^2$Department of Linguistics, \\
$^3$Department of Computer Science,\\
University of California San Diego \\
}

\abstract{
For many low-resource languages, the only available language models are large multilingual models trained on many languages simultaneously.
Despite state-of-the-art performance on reasoning tasks, we find that these models still struggle with basic grammatical text generation in many languages.
First, large multilingual models perform worse than bigrams for many languages (e.g. 24\% of languages in XGLM 4.5B; 43\% in BLOOM 7.1B) using FLORES perplexity as an evaluation metric.
Second, when we train small monolingual models with only 125M parameters on 1GB or less data for 350 languages, these small models outperform large multilingual models both in perplexity and on a massively multilingual grammaticality benchmark.
To facilitate future work on low-resource language modeling, we release \textit{Goldfish}, a suite of over 1,000 small monolingual language models trained comparably for 350 languages.
These models represent the first publicly-available monolingual language models for 215 of the languages included.
\\ \newline  
\begin{center}
   \raisebox{-0.5pt}{\includegraphics[scale=0.08]{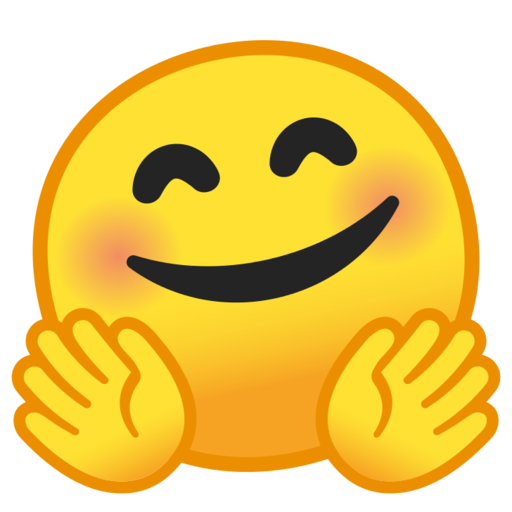}} \href{https://huggingface.co/goldfish-models}{~\texttt{Models and training data}} ~~~ \faGithub
 \href{https://github.com/tylerachang/goldfish}{~~\texttt{Training and evaluation code}} 
\end{center}
}

\begin{document}

\maketitleabstract

\section{Introduction}

Language modeling research in low-resource languages often relies on large multilingual models trained on many languages simultaneously \citep{conneau-etal-2020-unsupervised,adelani-etal-2021-masakhaner,ebrahimi2021americasnli,lin2022xglm,hangya2022improving,imanigooghari-etal-2023-glot500}.
For many low-resource languages, a dedicated model optimized for that language does not exist.
For example, in an analysis of publicly available models on Hugging Face, we find that 215 of the 350 languages in this paper did not have a monolingual text-generation model prior to the Goldfish models, and 47 of the languages did not have any text-generation models at all \citep{arnett2025analysis}. 

This lack of dedicated models hinders comparability of results across models and languages \citep{bandarkar2024belebele}, and it contributes to model under-performance in low-resource languages \citep{wu-dredze-2020-languages,blasi2022systematic}.
These barriers to research in low-resource languages are likely to exacerbate existing inequities across language communities in NLP research \citep{bender2011achieving,joshi-etal-2020-state}.

In particular, we find in this paper that large multilingual language models still struggle with basic next token prediction---a prerequisite for text generation---in low-resource languages.
To establish a baseline for basic text generation performance in low-resource languages, we introduce \textbf{Goldfish}\footnote{This name refers to shared qualities between our models and goldfish (\textit{Carassius auratus}); they are small, there are many of them, and they are known for their poor memories (perhaps inaccurately; \citealp{carey-2024-goldfish}). If an acronym is desired, Goldfish can stand for \textbf{G}enerative aut\textbf{O}regressive \textbf{L}ow-resource mo\textbf{D}els \textbf{F}or l\textbf{I}mited-compute \textbf{S}ystem \textbf{H}ardware.}, a suite of over 1000 monolingual language models for 350 diverse languages.

The Goldfish reach lower perplexities than XGLM \citep{lin2022xglm}, BLOOM 7.1B \citep{bigscience-bloom}, and MaLA-500 \citep{lin-2024-mala500} on 98 out of 204 FLORES languages, despite each Goldfish model being over 10$\times$ smaller (\S\ref{sec:flores-ppls}).
The Goldfish also outperform simple bigram models, which are surprisingly competitive with larger models for low-resource languages (e.g. lower perplexities than BLOOM 7.1B on 43\% of its languages; \S\ref{sec:flores-ppls}).
Finally, the Goldfish outperform large multilingual models on MultiBLiMP (\citealp{jumelet2025multiblimp10massivelymultilingual}; \S\ref{sec:reasoning}), a massively multilingual grammaticality benchmark.
However, despite better perplexities and grammaticality performance, the Goldfish perform at around chance, similar to other small multilingual models, on reasoning benchmarks (\S\ref{sec:reasoning}).

Thus, the primary contributions of this short paper are (1) to show that large multilingual models still suffer from poor next token prediction performance in low-resource languages, (2) to demonstrate that small monolingual models often exhibit lower perplexities and more grammatical next token predictions for such languages, and (3) to address the lack of available models in low-resource languages by releasing comparable monolingual models for 350 languages.
Models, training data, and code are available at: \url{https://huggingface.co/goldfish-models}.

\setlength{\belowcaptionskip}{-0.0cm}
\begin{figure*}[t]
    \centering
    \begin{minipage}[c]{0.5\textwidth}
        \centering
        \includegraphics[width=\linewidth]{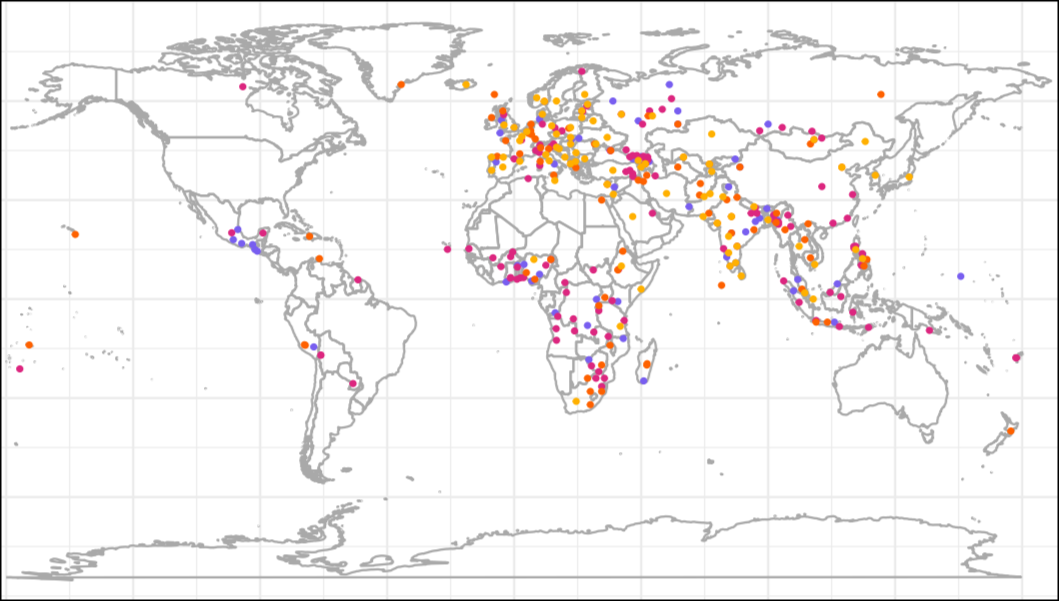}
    \end{minipage}
    \hspace{1em}
    \begin{minipage}[c]{0.45\textwidth}
        \centering
        \footnotesize
        \definecolor{fish5}{HTML}{785EF0}
        \definecolor{fish10}{HTML}{DC267F}
        \definecolor{fish100}{HTML}{FE6100}
        \definecolor{fish1000}{HTML}{FFB000}
        \renewcommand{\arraystretch}{1.3}
        \begin{tabular}{|p{1.5cm}|p{4.75cm}|}
            \hline
            \textbf{Data size} & \textbf{Model output} \\
            \hline
            \textcolor{fish5}{\textbf{5MB}} & {\ttfamily \textcolor{fish5}{Goldfish are a few years of the most of the most of the most}...} \\
            \textcolor{fish10}{\textbf{10MB}} & {\ttfamily \textcolor{fish10}{Goldfish are a great way to the best way to the best way}...} \\
            \textcolor{fish100}{\textbf{100MB}} & {\ttfamily \textcolor{fish100}{Goldfish are a great way to get your fish in the wild.}} \\
            \textcolor{fish1000}{\textbf{1GB}} & {\ttfamily \textcolor{fish1000}{Goldfish are a species of fish that are found in the sea.}} \\
            \hline
        \end{tabular}
    \end{minipage}
    \caption{Left: Map of the 350 languages for which Goldfish models are available, using coordinates from Glottolog \citep{hammarstrom-etal-2021-glottolog}. Color indicates the largest dataset size for that language. Right: Sample model outputs completing the prompt \texttt{``Goldfish are''} for the \texttt{eng\_latn} (English) model for each dataset size, using sampling temperature zero. Grammatical text generation begins to emerge in the 100MB-dataset model (available for 166 languages), but the lower-resource models still achieve better perplexities than previous models for many low-resource languages (\S\ref{sec:flores-ppls}).}
    \label{fig:goldfish-results}
\end{figure*}
\setlength{\belowcaptionskip}{0cm}

\section{Related Work}

Low resource language modeling often leverages multilingual pretraining, where a model is trained on multiple languages simultaneously \citep{pires2019multilingual,conneau-etal-2020-unsupervised}.
Indeed, this can improve low-resource performance, particularly when models have sufficient capacity and the multilingual data is from related or typologically similar languages \citep{kakwani2020indicnlpsuite,ogueji2021small,chang-etal-2023-multilinguality}.
However, monolingual models have still been shown to achieve better performance than multilingual models for many languages (e.g. \citealp{martin-etal-2020-camembert,pyysalo-etal-2021-wikibert,gutierrez2021maria,luukkonen-etal-2023-fingpt}).
Thus, it appears that existing multilingual language models are still limited by model capacity or limited data in low-resource languages \citep{conneau-etal-2020-unsupervised,chang-etal-2023-multilinguality}.

Notably, the training datasets for massively multilingual models are often heavily skewed towards high-resource languages.
For example, XGLM 4.5B is trained on over 7000$\times$ more Norwegian (71GB; 5.4M native speakers) than Quechua (0.01GB; 7.3M native speakers; \citealp{lin2022xglm,ethnologue-2024}).
In a more extreme case, BLOOM is trained on only 0.07MB of Akan (8.1M native speakers) out of 1.61TB total (\mbox{4e-6}\% of the pretraining dataset; \citealp{bigscience-bloom}).
These extremely small quantities of low-resource language data often do not leverage recent efforts to compile text data in low-resource languages \citep{costa2022no,imanigooghari-etal-2023-glot500,kudugunta-etal-2023-madlad400}, and the data imbalances are likely to severely hinder performance in low-resource languages.
Indeed, we find that these models have worse perplexities than simple bigram models for many languages (\S\ref{sec:flores-ppls}).
Unfortunately, comparable monolingual language models across many diverse languages have yet to be studied or released.

\section{Models and Datasets}
To compare to larger multilingual models, we introduce the Goldfish models, a suite of 1154 monolingual Transformer language models pretrained for 350 languages. The largest model for each language is 125M parameters.
We train models on 5MB, 10MB, 100MB, and 1GB of text when available after byte premium scaling \citep{arnett2024bit}.
Figure \ref{fig:goldfish-results} shows a geographic map of the 350 languages, with coordinates from Glottolog \citep{hammarstrom-etal-2021-glottolog}, along with sample outputs from the English model for each dataset size.

\subsection{Training Datasets}
\label{sec:datasets}
We merge the massively multilingual text datasets compiled in \citet{chang-etal-2023-multilinguality}, Glot500 \citep{imanigooghari-etal-2023-glot500}, and MADLAD-400 (\citealp{kudugunta-etal-2023-madlad400}) per language.
We deduplicate repeated sequences of 100 UTF-8 bytes and drop languages with only Bible data.
Full dataset details are in \S\ref{app:dataset-details}.
To facilitate fair evaluations, we hold out FLORES-200 and AmericasNLI from all datasets \citep{costa2022no,ebrahimi2021americasnli}. 
We conduct a contamination analysis to determine whether FLORES is contaminated in our datasets and find that for 98\% of languages, less than 10 out of 2000 FLORES sequences appear in the training dataset at all (\S\ref{app:contamination}).

To sample pretraining datasets of the desired sizes in a language $L$, we first use the Byte Premium Tool \citep{arnett2024bit} to estimate the \textbf{byte premium} for $L$, the number of UTF-8 bytes required to encode comparable text in $L$ relative to \texttt{eng\_latn} (English).
For example, \texttt{khm\_khmr} (Khmer) has byte premium 3.91, meaning that it uses approximately 3.91$\times$ as many UTF-8 bytes as English to encode content-matched text.
We divide each dataset size by the estimated byte premium for the corresponding language, thus measuring all datasets in units of ``equivalent'' English text bytes.
We sample datasets to train monolingual language models on 5MB, 10MB, 100MB, and 1GB when available after byte premium scaling.\footnote{The languages with 5MB-dataset models are a subset of the languages with 10MB-dataset models, and similarly for the 100MB and 1GB dataset sizes.}
These are equivalent to roughly 1M, 2M, 20M, and 200M tokens of English text respectively; including 10 epochs of repetition, the 1GB-dataset models are trained on the equivalent of roughly 2B English tokens.
When a 1GB dataset is not available for a language after byte premium scaling, we include a \textbf{full} model (267 languages) trained on the entire dataset in that language, for use cases that seek to maximize performance in a specific low-resource language.

As we cap our dataset sizes at 1GB per language, our models are trained on much less data than is otherwise available for high-resource languages. Still, when we compare our dataset sizes to FineWeb2 \citep{penedo2025fineweb} for the languages covered by both Goldfish and FineWeb2, 106 of those languages have more data in our dataset than in FineWeb2. We attribute this partially to our approach to data collection, which involves seeking out language-specific resources rather than relying solely on language identification models, which can be unreliable for web data in low-resource languages \citep{suarez2026commonlid}. We release our training datasets at: \url{https://huggingface.co/datasets/goldfish-models/fish-food}. 

\setlength{\belowcaptionskip}{-0.0cm}
\begin{table*}[t]
    \centering
    \footnotesize
    \renewcommand{\arraystretch}{1.3}
    \begin{tabular}{|p{2.75cm}|c|cccc|}
        \cline{1-6}
        Goldfish data size &
        \multicolumn{1}{c|}{\# Langs} &
        \multicolumn{1}{c|}{Goldfish} &
        \multicolumn{1}{c|}{Bigrams} &
        \multicolumn{1}{c|}{XGLM 4.5B} &
        \multicolumn{1}{c|}{MaLA-500 10B} \\ 
        \hline
        1000MB & 73 & \textbf{76.9} & 112.3 & 78.6 & 84.7 \\
        100MB & 22 & \textbf{102.7} & 132.6 & 143.9 & 121.7 \\
        10MB, 5MB & 5 & \textbf{130.5} & 148.3 & 183.1 & 135.0 \\
        \hline
    \end{tabular}
    \normalsize
    \caption{Mean FLORES perplexity ($\downarrow$) for the 100 languages in XGLM 4.5B, MaLA-500, and FLORES, separated by maximum Goldfish dataset size. The Goldfish languages are a strict superset of these languages.
    }
    \label{tab:mean-ppls}
\end{table*}
\setlength{\belowcaptionskip}{0cm}

\setlength{\belowcaptionskip}{-0.0cm}
\begin{table*}[t]
    \centering
    \footnotesize
    \renewcommand{\arraystretch}{1.3}
    \begin{tabular}{|p{2.25cm}|ccccc|}
        \cline{2-6}
        \multicolumn{1}{c|}{} &
        \multicolumn{1}{c|}{Bigrams} &
        \multicolumn{1}{c|}{XGLM 4.5B} &
        \multicolumn{1}{c|}{XGLM 7.5B} &
        \multicolumn{1}{c|}{BLOOM 7.1B} &
        \multicolumn{1}{c|}{MaLA-500 10B} \\ 
        \hline
        Bigrams & & 24 / 102 & 0 / 30 & 20 / 46 & 11 / 175 \\
        Goldfish (ours) & \textbf{202} / 202 & \textbf{60} / 102 & 2 / 30 & \textbf{32} / 46 & \textbf{111} / 175 \\
        \hline
    \end{tabular}
    \normalsize
    \caption{FLORES perplexity win rates for each row vs. column model. For example, Goldfish reach lower log-perplexities than MaLA-500 for 111/175 (63\%) of FLORES languages in both Goldfish and MaLA-500.}
    \label{tab:win-rates}
\end{table*}
\setlength{\belowcaptionskip}{0cm}

\subsection{Architectures and Pretraining}
\label{sec:model-training}

We train monolingual language models for five dataset sizes when available after byte premium scaling: \textbf{5MB}, \textbf{10MB}, \textbf{100MB}, \textbf{1GB}, and \textbf{full}.
The full dataset size (including all available data) is only included if a 1GB dataset is not available for a language.
In total, the Goldfish include 350 5MB-dataset models, 288 10MB-dataset models, 166 100MB-dataset models, 83 1GB-dataset models, and 267 full-dataset models (1154 models total).

For each language and each dataset size, we pretrain an autoregressive GPT-2 Transformer language model from scratch \citep{radford-etal-2019-language}.
For the 1GB, 100MB, and full dataset sizes, we use the 125M-parameter architecture equivalent to GPT-1 \citep{radford-etal-2018-improving}, which has a similar parameter count to BERT-base and RoBERTa \citep{devlin-etal-2019-bert,liu-etal-2019-roberta}.
Because larger models do not appear to outperform smaller models for very small datasets \citep{chang-etal-2023-multilinguality}, we use the small model size (39M parameters) from \citet{turc2019well} for the 10MB and 5MB dataset sizes.
Full hyperparameters are reported in \S\ref{app:pretraining-details}.

Each model has a custom monolingual tokenizer, which is trained with a vocabulary size of 50K \citep{liu-etal-2019-roberta} on the same dataset size as their corresponding model (including byte premium scaling).
We use Unigram tokenizers trained with the SentencePiece implementation \citep{kudo-richardson-2018-sentencepiece}. Training text is randomly sampled from the dataset for the desired language.\footnote{To avoid memory errors, we limit tokenizer training text to 100MB after byte premium scaling.}
After tokenizer training, we tokenize each training dataset, concatenating text lines such that each sequence contains exactly 512 tokens.
We run tokenization before shuffling and sampling to the desired dataset sizes, so our sequences of 512 tokens preserve contiguous text where possible, although several of our source corpora only exist in shuffled form.
Finally, we sample our tokenized datasets to 5MB, 10MB, 100MB, and 1GB after byte premium scaling.\footnote{When de-tokenized, the tokenized datasets are slightly smaller than the original text datasets, because the tokenizer truncates lines to create 512-token sequences. Reported dataset sizes account for truncation.}

We train each model for 10 epochs of the training data; multiple epochs of pretraining is beneficial in data-constrained scenarios \citep{muennighoff-etal-2023-scaling}, but pretraining on more than 10 epochs often leads to overfitting (increases in eval loss) in the 5MB scenarios.
All language model pretraining runs together take a total of $1.65 \times 10^{20}$ FLOPs.
This is less than $1/1900\times$ the computation used to train the original 175B-parameter GPT-3 model (\citealp{brown-etal-2020-language}; $3.14 \times 10^{23}$ FLOPs). Further pretraining details are reported in \S\ref{app:pretraining-details}.

\subsection{Training Bigram Baselines}

Because there do not exist other models for many of these languages, we also train bigram baseline models on the same training data and with the same tokenizer for each language.
Each bigram model computes the probability of each token $w_i$ as $P(w_i | w_{i-1})$, computed based on raw bigram counts in the tokenized Goldfish dataset.
When a bigram is not observed in the dataset, we use backoff to unigram probability with a penalty multiplier of $\lambda=0.40$ (i.e. ``stupid backoff''; \citealp{brants-etal-2007-large}).
We do not consider $n$-grams for $n>2$ because those $n$-grams often resort to backoff and are therefore much more sensitive to the backoff penalty term $\lambda$.

\section{Evaluations}

\subsection{FLORES Log-Perplexity}
\label{sec:flores-ppls}

We first evaluate our models on FLORES-200 log-perplexity \citep{costa2022no} (equivalently, negative log-likelihood; \citealp{lin-2024-mala500}).
To avoid confounds from different tokenizers across models, we compute log-perplexities at the sequence level.
Specifically, regardless of its tokenization, a language model $\mathcal{M}$ assigns some probability $P_{\mathcal{M}}(s)$ to each sequence $s$ in FLORES.
In most cases, $s$ is a single sentence.
For fair comparison with multilingual models that need to determine the input language during the early parts of a sequence, we compute log-perplexity of the second half $s_1$ of each sequence given the first half $s_0$.
We then compute the mean over sequences:
\setlength{\abovedisplayskip}{0.2cm}
\setlength{\belowdisplayskip}{0.2cm}
\begin{equation}
\label{eq:ppl}
\textrm{LogPPL}_{\mathcal{M}} = \textrm{mean}_{s} \Big( - \textrm{log}( P_{\mathcal{M}}(s_1 | s_0)) \Big)
\end{equation}
A lower log-perplexity indicates better performance, where $\mathcal{M}$ assigns higher probabilities to ground truth text (FLORES sequences).
While imperfect, perplexity does not require annotated text data, it is predictive of performance on a variety of downstream tasks 
\citep{xia-etal-2023-training}, and it has been used to measure language model quality in previous work \citep{kaplan-etal-2020-scaling,hoffmann-etal-2022-training,lin-2024-mala500}.

We compare the Goldfish with XGLM 4.5B (\citealp{lin2022xglm}; 134 languages), XGLM 7.5B (30 languages), BLOOM 7.1B (\citealp{bigscience-bloom}; 46 languages), and MaLA-500 10B (\citealp{lin-2024-mala500}; 534 languages).
We also compare to our bigram baselines, which were trained on the same datasets as the Goldfish models.
In all cases, we use the Goldfish model trained on the maximum amount of data in a language (maximum 1GB).

The Goldfish reach lower log-perplexities than all four comparison models on 98 of the 204 FLORES languages.
On average, the Goldfish reach 13\% lower perplexities than XGLM 4.5B, and 11\% lower than MaLA-500 10B (Table~\ref{tab:mean-ppls}).
To ensure that these results are not driven by a small subset of specific languages, in Table~\ref{tab:win-rates} we also report the pairwise ``win'' rates for Goldfish and bigrams vs. all four comparison models, for the set of FLORES languages shared between each pair.
The Goldfish models have a perplexity win rate above 50\% against all comparison models except XGLM 7.5B, which considers only 30 fairly high-resource languages \citep{lin2022xglm}.
Notably, the \textit{bigram} models also reach lower perplexities than large multilingual models for a nontrivial number of languages: 24\% of languages in XGLM 4.5B and 43\% of languages in BLOOM 7.1B.
Still, the bigrams have worse perplexities than Goldfish for all languages.
Perplexities for individual languages and models are available in our GitHub.

\begin{table*}[ht]
\centering
\small
\renewcommand{\arraystretch}{1.1}
\begin{tabular}{|r|r|rrrrrrr|}
\hline
Dataset / Model & Chance & Goldfish & Gemma 3 & BLOOM & XGLM & Gemma 3 & LLaMA 3 & XGLM \\
\textbf{} & \textbf{} & 124M & 270M & 560M & 564M & 1B & 1B & 1.7B \\ \hline
MultiBLiMP (avg) & 50.0 & \textbf{78.8} & 72.3 & 64.2 & 66.6 & 78.0 & 77.4 & 62.7 \\ 
Belebele (avg) & 25.0 & 28.2 & 22.7 & 28.9 & \textbf{29.0} & 22.7 & 28.9 & 28.2 \\
XCOPA (avg) & 50.0 & 55.1 & 54.6 & 54.4 & 55.2 & \textbf{59.7} & 55.9 & 57.6 \\
XStoryCloze (avg) & 50.0 & 52.3 & 52.7 & 52.6 & 53.1 & \textbf{58.5} & 55.1 & 56.2 \\
\hline
\end{tabular}
\caption{Average performance on downstream benchmarks for all models tested.}
\label{tab:multilingual}
\end{table*}

\subsection{Downstream Tasks}
\label{sec:reasoning}

Next, we evaluate linguistic knowledge (grammaticality) with MultiBLiMP \citep{jumelet2025multiblimp10massivelymultilingual} as implemented in the LM Evaluation Harness \citep{eval-harness}; this covers 74 of the Goldfish languages.
We compare against popular small multilingual models: 
BLOOM 560M \citep{bigscience-bloom},
XGLM (564M and 1.7B; \citealp[]{lin2021few}),
Gemma 3 (270M and 1B base models; \citealp[]{gemmateam2025gemma3technicalreport}), and
Llama 3.2 (1B base model; \citealp[]{meta2024llama3}). 
Results are in Table~\ref{tab:multilingual}.
The Goldfish models have higher average accuracy than any of the multilingual models, and they have the highest accuracy of all models tested for 25 of the 74 languages.
This result highlights the benefits of small monolingual models, especially for languages which only account for a small portion of the training data in multilingual models.

We also evaluate the same models on three popular multilingual reasoning benchmarks:  Belebele (121 languages, reading comprehension; \citealp{bandarkar2024belebele}), XCOPA (11 languages, commonsense reasoning; 
\citealp{ponti-etal-2020-xcopa}), and XStoryCloze (10 languages, story commonsense reasoning; \citealp{lin2022xglm}). 
All models are evaluated zero shot with log-probability solution ranking, with no fine-tuning or instruction tuning.
Unfortunately, all models perform quite poorly (close to chance accuracy; Table~\ref{tab:multilingual}), suggesting that available multilingual reasoning evaluations are not appropriate for pretrained-only models of this scale.

\section{Discussion}

Our results demonstrate that large multilingual language models still struggle with basic grammatical text generation for many languages (e.g. often worse perplexities than bigrams).
The Goldfish (small monolingual models) exhibit lower perplexities and more grammatical next token predictions for many low-resource languages.
These results make the Goldfish suitable baselines for basic grammatical text generation in diverse languages, motivating future work developing low-resource language models.
Furthermore, the Goldfish are accessible to labs with limited compute budgets, they are trained on a roughly human-like amount of data \citep{warstadt-etal-2023-findings}, and they are trained to be maximally comparable across languages.
Future work may investigate precisely when larger-scale multilingual pretraining provides benefits to lower-resource languages; for example, it may be that abstract reasoning patterns and heuristics are often more language-agnostic than grammatical text generation, and thus larger-scale multilingual pretraining primarily benefits the former.

\section{Conclusion}
In this paper, we pretrain and release Goldfish, a suite of over 1000 monolingual language models for 350 languages. For the majority of these languages, the Goldfish represent the first monolingual model dedicated for that language.
The Goldfish achieve perplexities that are competitive with, and on average lower than, state-of-the-art multilingual language models across languages.
However, similar to multilingual models of the same scale, the Goldfish still struggle with reasoning tasks. This suggests that smaller monolingual models may better represent linguistic knowledge of the target language, but they do not perform well on more complex tasks at this scale.
We publicly release the Goldfish models to promote future research pushing the limits of language model capabilities in low-resource languages.

\section*{Limitations}

\paragraph{Comparability and availability.}
In order to include as many low-resource languages as possible, the Goldfish models are trained on corpora compiled from a wide variety sources (\S\ref{app:dataset-details}).
Still, 5MB of text (roughly 1M tokens) is not publicly available for many of the world's languages.
Even where text is available, corpora for different languages vary significantly both in cleanliness and domain coverage (e.g. news vs. social media vs. books).
Thus, while we release models trained on comparable quantities of text in different languages (including accounting for byte premiums; \citealp{arnett2024bit}; \S\ref{sec:datasets}), the models are not perfectly comparable across languages.
In fact, it is likely that such perfect comparability is impossible given the diversity of the world's languages, cultures, and language use.
Even directly translated datasets are not perfectly comparable across languages \citep{untranslatability}.
Thus, the Goldfish models aim to maximize model and dataset comparability across languages while still covering a wide variety of languages.

\paragraph{Monolinguality.}
By design, all of the Goldfish models are monolingual.
For low-resource languages, training on closely related languages would likely improve performance \citep{conneau-etal-2020-unsupervised,chang-etal-2023-multilinguality}.
However, adding multilingual data introduces concerns such as the choice of added languages (some languages have more closely related languages in our dataset than others), quantities of added data, and model capacity limitations.
To maximize comparability across languages and to allow the models to serve as clearly-defined baselines, we train all Goldfish models monolingually.
Of course, language-annotated text datasets inevitably contain mislabeled text, particularly for similar languages \citep{caswell-etal-2020-language,blevins-zettlemoyer-2022-language,kreutzer-etal-2022-quality}.
Thus, we cannot guarantee that our models are entirely free from cross-language contamination, although they are monolingual to the best ability of current language identification models.

\paragraph{Model and dataset sizes.}
Because the Goldfish are focused on low-resource languages, we restrict all models to 1GB of training text (after byte premium scaling; \citealp{arnett2024bit}).
For the majority of the world's languages, 1GB is sufficient to include all publicly available text data in the language.
At these small dataset sizes, larger models do not appear to provide significant benefit over smaller models \citep{kaplan-etal-2020-scaling,hoffmann-etal-2022-training,chang-etal-2023-multilinguality}.
Thus, the largest Goldfish model that we train for each language has 125M parameters and is trained on a maximum of 1GB of text.
This is the same model size as GPT-1 \citep{radford-etal-2018-improving} or BERT \citep{devlin-etal-2019-bert}, and the 1GB dataset size is approximately 20\% of the dataset size of GPT-1 \citep{radford-etal-2018-improving}.

\paragraph{Downstream tasks.}
We evaluate the Goldfish models on FLORES log-perplexity (\S\ref{sec:flores-ppls}) and four multilingual benchmarks, including three reasoning benchmarks (\S\ref{sec:reasoning}). 
Perplexity is the only evaluation available for autoregressive language models in many languages before instruction-tuning and RLHF, but it has significant limitations.
Perplexity is not necessarily predictive of more specific capabilities \citep{hu-etal-2020-systematic,levy2024tasktokensimpactinput}, although it still provides reasonable signal for model performance \citep{xia-etal-2023-training}.
On the other hand, reasoning benchmarks require annotated datasets and thus often cover fewer languages.
Belebele (121 non-English languages; \citealp{bandarkar2024belebele}) is an exception, but even state-of-the-art models perform quite poorly on Belebele without instruction-tuning or few-shot prompting (\S\ref{sec:reasoning}).
Thus, our evaluations of model reasoning are not conclusive; we may primarily be measuring heuristics that allow the models to perform only somewhat above chance (arguably, this might still be considered a basic form of ``reasoning'').

Outside of reasoning benchmarks, we also evaluate on MultiBLiMP \citep{jumelet2025multiblimp10massivelymultilingual}, which evaluates linguistic knowledge. MultiBLiMP significantly expands the language coverage of multilingual grammaticality benchmarks; however, MultiBLiMP evaluates a very narrow aspect of grammatical knowledge (two types of subject-verb agreement).
We hope that tractable evaluation datasets with broad language coverage will become increasingly available in the future to enable more informative evaluation of models in a broader range of languages.

\paragraph{Risks and dataset licensing.}
Trained on a maximum of 1GB of text each, the Goldfish models have very limited capabilities relative to modern language models in high-resource languages.
The Goldfish are trained on publicly-released corpora used in previous NLP research (\S\ref{app:dataset-details}), but we cannot guarantee that the data is free from offensive content or personally identifying information.
Our models are small, which reduces the likelihood that they will regurgitate memorized text \citep{carlini2023quantifyingmemorizationneurallanguage}.
As far as we are aware, we do not include any datasets that prohibit use for language model training.
We report all included datasets in \S\ref{app:dataset-details}.
We will remove models for affected languages if contacted by dataset owners.

\section*{Acknowledgements}
We would like to thank the other members of the UCSD Language and Cognition Lab for valuable discussion.
Z. Tu is supported by NSF award IIS-2127544 and NSF award IIS-2433768. 

\nocite{*}
\section{Bibliographical References}\label{sec:reference}
\bibliographystyle{lrec2026-natbib}
\bibliography{lrec2026-example}

\clearpage
\appendix

\section{Appendix}
\label{sec:appendix}

\subsection{Training Dataset Details}
\label{app:dataset-details}

\paragraph{Data sources.}
As described in \S\ref{sec:datasets}, we merge the text datasets compiled in \citet{chang-etal-2023-multilinguality}, Glot500 \citep{imanigooghari-etal-2023-glot500}, and MADLAD-400 (clean split; \citealp{kudugunta-etal-2023-madlad400}).
These datasets include popular multilingual corpora such as OSCAR \citep{suarez2019asynchronous,AbadjiOrtizSuarezRomaryetal.2021}, Wikipedia \citep{wikipedia}, No Language Left Behind \citep{costa2022no}, and others.
Together, these datasets take advantage of both automatically crawled datasets with automated language identification and targeted datasets manually annotated for specific low-resource languages.
All included datasets are publicly available; see Limitations for licensing concerns.
Comprehensively, the Goldfish dataset includes:
\begin{itemize}[leftmargin=0.5cm,itemsep=0.1cm,topsep=0.25cm]
\item \citet{chang-etal-2023-multilinguality}:\newline
OSCAR \citep{suarez2019asynchronous,AbadjiOrtizSuarezRomaryetal.2021}, Wikipedia \citep{wikipedia}, No Language Left Behind \citep{costa2022no}, Leipzig Corpora Collection \citep{goldhahn2012building}, eBible translations \citep{eBible}, Tatoeba \citep{tiedemann2012parallel,tiedemann-2020-tatoeba}, AfriBERTa \citep{ogueji2021small},  NusaX \citep{winata2022nusax}, AmericasNLP \citep{mager2021findings}, Nunavut Hansard Inuktitut–English Parallel Corpus \citep{joanis2020inuktitut}, Cherokee-English ChrEn dataset \citep{zhang2020chren}, Cherokee Corpus \citep{cherokee_dict}, Cree Corpus \citep{teodorescu-etal-2022-cree}, Languages of Russia \citep{zaydelman2016}, Evenki Life newspaper \citep{zueva2020finite}, transcribed Fula Speech Corpora \citep{fula2023}, IsiXhosa \citep{isixhosa_ner_corpus}, Ewe Language Corpus \citep{gelr2021}, Makerere Luganda Corpora \citep{mukiibi2022makerere}, CMU Haitian Creole dataset \citep{haitian_cmu}, Tigrinya Language Modeling Dataset \citep{gaim2021monolingual}, and Ulukau \citep{ulukau}.
\item Glot500 \citep{imanigooghari-etal-2023-glot500}:\newline
AI4Bharat \citep{AI4Bharat},
AI FOR THAI LotusCorpus \citep{aiforthai},
Arabic Dialects Dataset \citep{el-haj-etal-2018-arabic},
AfriBERTa \citep{ogueji2021small},
AfroMAFT \citep{adelani-etal-2022-thousand,xue-etal-2021-mt5},
Anuvaad \citep{Anuvaad},
AraBench \citep{sajjad-etal-2020-arabench},
Autshumato \citep{Autshumato}
Bloom Library \citep{DBLP:conf/emnlp/LeongNMFOW22}, 
CC100 \citep{conneau-etal-2020-unsupervised}, 
CCNet \citep{wenzek-etal-2020-ccnet}, 
CMU Haitian Creole \citep{haitian_cmu},
SADiLaR NCHLT corpus \citep{sadilar},
Clarin \citep{clarin},
DART \citep{alsarsour-etal-2018-dart},
Earthlings \citep{DBLP:journals/lre/Dunn20}, 
FFR Dataset \citep{ffr},
GiossaMedia \citep{gongora-etal-2022-use, gongora-etal-2021-experiments},
Glosses \citep{camacho-collados-etal-2016-large},
Habibi \citep{el-haj-2020-habibi},
HinDialect \citep{bafna2022empirical}, 
HornMT \citep{HornMT},
IITB \citep{kunchukuttan-etal-2018-iit},
IndicNLP \citep{nakazawa-etal-2021-overview},
Indiccorp \citep{kakwani2020indicnlpsuite}, 
isiZulu \citep{isizulu},
JParaCrawl \citep{morishita-etal-2020-jparacrawl}, 
kinyarwandaSMT \citep{KinyaSMT},
LeipzigData \citep{goldhahn2012building}, 
LINDAT \citep{lindat},
Lingala Song Lyrics \citep{lingala-songs},
LyricsTranslate \citep{lyrics-translate},
mC4 \citep{raffel-etal-2020-exploring}, 
MTData \citep{gowda-etal-2021-many}, 
MaCoCu \citep{DBLP:conf/eamt/BanonEFGKLNSRRS22}, 
Makerere MT Corpus \citep{mukiibi2022makerere},
Masakhane Community \citep{masakhane},
Mburisano Covid Corpus \citep{mburisano},
Menyo20K \citep{adelani-etal-2021-effect},
Minangkabau corpora \citep{koto-koto-2020-towards}, 
MoT \citep{palen-michel-etal-2022-multilingual},
NLLB seed \citep{costa2022no},
Nart Abkhaz text \citep{abkhaz},
OPUS \citep{tiedemann2012parallel}, 
OSCAR \citep{suarez2019asynchronous}, 
ParaCrawl \citep{banon-etal-2020-paracrawl},
Parallel Corpora for Ethiopian Languages \citep{teferra-abate-etal-2018-parallel}, 
Phontron \citep{neubig11kftt},
QADI \citep{abdelali-etal-2021-qadi},
Quechua-IIC \citep{zevallos2022introducing}, 
SLI GalWeb.1.0 \citep{agerri-etal-2018-developing},
Shami \citep{abu-kwaik-etal-2018-shami},
Stanford NLP \citep{stanford-nlp},
StatMT \citep{statmt},
TICO \citep{anastasopoulos-etal-2020-tico},
TIL \citep{mirzakhalov-etal-2021-large},
Tatoeba \citep{tiedemann-2020-tatoeba},
TeDDi \citep{moran-etal-2022-teddi},
Tilde \citep{rozis-skadins-2017-tilde},
W2C \citep{11858/00-097C-0000-0022-6133-9}, 
WAT \citep{nakazawa-etal-2022-overview},
WikiMatrix \citep{schwenk-etal-2021-wikimatrix}, 
Wikipedia \citep{wikipedia},
Workshop on NER for South and South East Asian Languages \citep{singh-2008-named}, and XLSum \citep{xlsum}.
\item MADLAD-400 \citep{kudugunta-etal-2023-madlad400}:\newline
CommonCrawl \citep{commoncrawl-2022}.
\end{itemize}
We start with the corpus from \citet{chang-etal-2023-multilinguality}.
We then merge the dataset per language with Glot500 for languages that have not yet reached our 1GB maximum (after byte premium scaling).
Then, we merge the dataset with MADLAD-400 for languages that have still not reached our 1GB maximum.
We also add MADLAD-400 for languages with short average line lengths (less than 25.0 tokens), to make use of MADLAD-400's longer contiguous sequences.
To allow comparisons on popular low-resource language evaluations, we exclude FLORES-200 \citep{costa2022no} and AmericasNLI \citep{ebrahimi2021americasnli} from all dataset merging.
For each dataset, we exclude languages that contain only Bible data.
Because there is likely significant overlap between different dataset sources, we deduplicate repeated sequences of 100 UTF-8 bytes for each language \citep{lee2021deduplicating}.

\paragraph{Language codes.}
To enable dataset merging per language, several datasets must be converted to ISO 639-3 language codes and ISO 15924 script codes.
In some cases, this introduces ambiguity because datasets can be labeled as individual language codes (e.g. \texttt{quy\_latn} for Ayacucho Quechua and \texttt{quz\_latn} for Cusco Quechua) or as macrolanguage codes (e.g. \texttt{que\_latn} for Quechua).
In these cases, we compile both a macrolanguage dataset and individual language datasets.
Datasets labeled with individual codes contribute both to their individual dataset and their umbrella macrolanguage dataset; datasets labeled with macrolanguage codes contribute only to the macrolanguage dataset.
For example, we have individual \texttt{quy\_latn} and \texttt{quz\_latn} datasets, both of which contribute to a larger \texttt{que\_latn} dataset, which also contains datasets labeled only with \texttt{que\_latn}.
These ambiguities primarily appear for lower-resource languages.

We also drop several redundant language codes:
\begin{itemize}[leftmargin=0.5cm,itemsep=0.0cm,topsep=0.1cm]
\item We drop \texttt{ory\_orya} (Odia) in favor of the macrocode \texttt{ori\_orya} because \texttt{ory\_orya} is the only individual language within \texttt{ori\_orya} for which we have any data.
\item For the same reason, we drop \texttt{npi\_deva} (Nepali) in favor of the macrocode \texttt{nep\_deva}.
\item For the same reason, we drop \texttt{swh\_latn} (Swahili) in favor of the macrocode \texttt{swa\_latn}.
\item We drop \texttt{cmn\_hans} (Mandarin) in favor of the macrocode \texttt{zho\_hans} (Chinese) because the \texttt{zho\_hans} data is almost entirely in Mandarin. While less specific, \texttt{zho\_hans} is commonly used by other datasets.
For other Chinese languages, see their individual codes (e.g. \texttt{yue\_hant} for Cantonese).
We note that the similar code \texttt{zho\_hant} (traditional characters) is not primarily Mandarin.
\item We drop \texttt{hbs\_cyrl} and \texttt{hbs\_latn} (Serbo-Croatian) because we have the individual languages Serbian (\texttt{srp\_cyrl} and \texttt{srp\_latn}), Croatian (\texttt{hrv\_latn}), and Bosnian (\texttt{bos\_cyrl} and \texttt{bos\_latn}).
\item We drop the deprecated code \texttt{ajp\_arab} (Levantine Arabic) in favor of \texttt{apc\_arab}.
\item We drop \texttt{ber\_latn} (Berber) because it is a collective code for distinct (and often not mutually intelligible) languages. We keep the constituent individual languages.
\item We drop \texttt{nah\_latn} (Nahuatl) because it is a collective code for distinct languages. We keep the constituent individual languages.
\end{itemize}
After merging, we have a dataset of 547GB of text covering 523 language-script combinations (486 unique language codes, 32 unique script codes).

\paragraph{Byte premiums.}
As described in \S\ref{sec:datasets}, we then scale our dataset sizes by estimated byte premiums \citep{arnett2024bit}.
A byte premium $b$ for a language $L$ indicates that content-matched (i.e. parallel) text in $L$ takes $b\times$ as many UTF-8 bytes to encode as English.
We use the Byte Premium Tool \citep{arnett2024bit} to compute or estimate the byte premium for all of our languages.
Byte premiums are pre-computed in the tool for high-resource languages.
For each novel low-resource language $L$, we use the tool (which uses a linear regression) to predict the byte premium for $L$ based on the character entropy for text in $L$ and the script type for $L$ (alphabet, abjad, abugida, or logography), as recommended for low-resource languages in \citet{arnett2024bit}.
Then, we have an estimated byte premium for every language in our dataset.
We clip each byte premium to a minimum of 0.70 and a maximum of 5.00; clipping occurs for only three languages (\texttt{lzh\_hant}, \texttt{wuu\_hani} $\to$ 0.70, \texttt{mya\_mymr} $\to$ 5.00).
As described in \S\ref{sec:datasets}, all of our training datasets (both for tokenizers and for the models themselves) are sampled based on size in bytes after byte premium scaling.
We drop languages with less than 5MB of text after byte premium scaling.

\paragraph{Dataset statistics.}
The resulting 350 Goldfish languages cover five continents, 28 top-level language families \citep{hammarstrom-etal-2021-glottolog}, and 32 scripts (writing systems).
All languages for which Goldfish models are available are listed in Table~\ref{tab:lang-list}.
We include the language name, ISO 639-3 language code, ISO 15924 script code, estimated byte premium, dataset size after byte premium scaling, dataset size in tokens, and proportion of the dataset from each of our four largest sources.
Raw dataset sizes before byte premium scaling can be obtained by multiplying the dataset size after byte premium scaling by the estimated byte premium.
Source dataset proportions are reported before deduplication.
The reported dataset sizes reflect the dataset for the Goldfish model trained on the maximum amount of data for that language (the \textbf{1GB}-dataset Goldfish when available, otherwise the \textbf{full}-dataset Goldfish).
Reported token counts use the tokenizer for the largest Goldfish model for that language.
Datasets can be downloaded at: \url{https://huggingface.co/datasets/goldfish-models/fish-food}.

\subsection{Contamination Analysis} \label{app:contamination}

While we aim to exclude the FLORES dataset from our training datasets, there is still the possibility that FLORES could be contaminated inadvertently due to the presence of web data.
To check for any potential contamination of FLORES in the datasets, we tokenize each FLORES sequence and the entire training dataset for each language that is in both Goldfish and FLORES (204 languages).
We compute the total number of times that the first 10 tokens of any FLORES example appears in the training dataset for the language.
For 72\% of languages, no FLORES examples appear in the training dataset at all.
For 98\% of languages, less than 10 FLORES examples appear in the training dataset (out of over 2000 FLORES examples total).
The only two languages with notable FLORES contamination are \texttt{smo\_latn} (Samoan; 7155 occurrences of FLORES examples in the training dataset) and \texttt{knc\_arab} (Central Kanuri; 371 occurrences of FLORES examples in the training dataset).

\subsection{Pretraining Details}
\label{app:pretraining-details}

\setlength\tabcolsep{3pt}
\begin{table}[t]
    \centering
    \small
    \renewcommand{\arraystretch}{1.11}
    \begin{tabular}{|>{\raggedright}p{3.1cm}|r|r|}
    \hline
    \textbf{Hyperparameter} & \textbf{5MB, 10MB} & \textbf{full, 100MB+} \\
    \hline
    Total parameters & 39M & 125M \\
    Layers & 4 & 12 \\
    Embedding size & 512 & 768 \\
    Hidden size & 512 & 768 \\
    Intermediate hidden & 2048 & 3072 \\
    Attention heads & 8 & 12 \\
    Attention head size & 64 & 64 \\
    \hline
    Learning rate & \multicolumn{2}{r|}{1e-4} \\
    Batch size & \multicolumn{2}{r|}{5MB: 4, 10MB: 8,} \\
    & \multicolumn{2}{r|}{100MB: 32, 1GB: 64} \\
    Epochs & \multicolumn{2}{r|}{10} \\
    Activation function & \multicolumn{2}{r|}{GELU} \\
    Max sequence length & \multicolumn{2}{r|}{512} \\
    Position embedding & \multicolumn{2}{r|}{Absolute} \\
    Learning rate decay & \multicolumn{2}{r|}{Linear} \\
    Warmup steps & \multicolumn{2}{r|}{10\% of pretraining} \\
    Adam $\epsilon$ & \multicolumn{2}{r|}{1e-6} \\
    Adam $\beta_1$ & \multicolumn{2}{r|}{0.9} \\
    Adam $\beta_2$ & \multicolumn{2}{r|}{0.999} \\
    Dropout & \multicolumn{2}{r|}{0.1} \\
    Attention dropout & \multicolumn{2}{r|}{0.1} \\
    \hline
    \end{tabular}
    \caption{Pretraining hyperparameters for Goldfish trained on different dataset sizes \citep{devlin-etal-2019-bert,turc2019well,radford-etal-2018-improving}.}
    \label{tab:hyperparameters}
\end{table}

\paragraph{Architectures.}
All of our models use the GPT-2 architecture \citep{radford-etal-2019-language}, changing only the number of layers, attention heads, and embedding sizes as in \citet{turc2019well}.
For the 100MB-, 1GB-, and full-dataset models, we use the 125M-parameter architecture equivalent to GPT-1 \citep{radford-etal-2018-improving} (similar to BERT-base and RoBERTa; \citealp{devlin-etal-2019-bert,liu-etal-2019-roberta}).
Because smaller models perform similarly to larger models in low-resource scenarios \citep{chang-etal-2023-multilinguality}, we use the small model size (39M parameters) from \citet{turc2019well} for the 10MB and 5MB dataset sizes.

\paragraph{Training hyperparameters.}
Language models are pretrained using the Hugging Face Transformers library \citep{wolf-etal-2020-transformers} and code from \citet{chang-bergen-2022-word}.
We refrain from extensive hyperparameter tuning to avoid biasing our hyperparameters towards English (or any other selected tuning language).
Instead, we adopt hyperparameters from previous work with minimal modifications.
To match the setup of our models and to prevent overfitting, we select hyperparameters based on models with fairly small training datasets relative to modern standards.
Specifically, following BERT \citep{devlin-etal-2019-bert}, we use learning rate 1e-4 for the 125M-parameter models (the same as RoBERTa for small batch sizes; \citealp{liu-etal-2019-roberta}; GPT-1 uses learning rate 2.5e-4; \citealp{radford-etal-2018-improving}).
Based on initial results using randomly-sampled languages, we find that learning rate 1e-4 also works well for the 39M-parameter models; this is in line with \citet{chang-etal-2023-multilinguality}, who find that learning rate 2e-4 works well for small models, and smaller learning rates reduce the speed of any potential overfitting.

Following GPT-1 (most similar to our models; \citealp{radford-etal-2018-improving}), we use batch size 64 (64$\times$512 = 32K tokens) for the 1GB-dataset models.
We find that these larger batch sizes lead to overfitting for small datasets, so we use batch sizes 4, 8, and 32 for \mbox{5MB-,} \mbox{10MB-,} and 100MB-dataset models respectively (determined based on initial experiments with randomly-sampled languages).
These correspond to batches of 2K, 4K, or 16K tokens.
For full-dataset models, we use the batch size that would be used if rounding the dataset size down to 5MB, 10MB, or 100MB (recall that we do not train a full-dataset model when the 1GB dataset is available for a language).

Models are each trained on one NVIDIA GeForce GTX TITAN X, GeForce RTX 2080 Ti, TITAN Xp, Quadro P6000, RTX A4500, RTX A5000, or RTX A6000 GPU.
In total, Goldfish pretraining takes the equivalent of approximately 15600 A6000 GPU hours.
Inference for FLORES perplexities and reasoning benchmarks takes approximately 250 A6000 GPU hours (primarily due to the large multilingual models used for comparison).
Dataset merging, deduplication, and tokenization takes approximately 1600 CPU core hours.

\subsection{FLORES Evaluation Details}
\label{app:flores-details}

In \S\ref{sec:flores-ppls}, we evaluate the Goldfish models, XGLM 4.5B, XGLM 7.5B, BLOOM 7.1B, MaLA-500 10B, and bigram models on FLORES log-perplexity (negative log-likelihood).
To avoid confounds from different tokenizers across models, we compute log-perplexities at the sequence level.
For fair comparison with multilingual models that need to determine the input language during the early parts of a sequence, we compute log-perplexity of the second half $s_1$ of each sequence given the first half $s_0$.
We then compute the mean over sequences:
\setlength{\abovedisplayskip}{0.2cm}
\setlength{\belowdisplayskip}{0.2cm}
\begin{equation}
\label{app_eq:ppl}
\textrm{LogPPL}_{\mathcal{M}} = \textrm{mean}_{s} \Big( - \textrm{log}( P_{\mathcal{M}}(s_1 | s_0)) \Big)
\end{equation}
A lower log-perplexity indicates better performance, where $\mathcal{M}$ assigns higher probabilities to ground truth text (FLORES sequences).
While imperfect, perplexity does not require annotated text data, it is predictive of performance on a variety of downstream tasks 
\citep{xia-etal-2023-training}, and it has been used to measure language model quality in previous work \citep{kaplan-etal-2020-scaling,hoffmann-etal-2022-training,lin-2024-mala500}.

In detail, the first and second half of each sequence are determined based on number of characters, so the halfway split is the same for all models considered.
We round to the nearest token when the halfway split is in the middle of a subword token.
Each model $\mathcal{M}$ then assigns some probability $P_{\mathcal{M}}(s_1 | s_0)$ regardless of tokenization, except for rounding the halfway point to the nearest token.
The probability for any [UNK] (unknown) token is set to random chance $1/v$ where $v$ is the tokenizer vocabulary size.\footnote{Otherwise, for unseen writing systems (e.g. Tibetan script \texttt{tibt} in XGLM), the probability $P(\textrm{[UNK]} | \textrm{[UNK] [UNK] ...})$ is very high, resulting in artificially low perplexities. Setting the [UNK] token probabilities to random chance has very little effect on log-perplexity scores except for the scenario of an unseen writing system.}
As our final log-perplexity score, we compute the mean negative-log-probability over all FLORES sequences in the target language.
Because perplexities generally use geometric means, we use arithmetic means for log-perplexities. 
The final equation is presented in Equation~\ref{app_eq:ppl}.

The mean FLORES perplexities for each model are reported in Table~\ref{tab:mean-ppls}.
For Goldfish models, we report the perplexity for the model trained on the largest dataset for the language (i.e. the 1GB-dataset model when available, otherwise the full-dataset model).
Perplexities per language for the 5MB-, 10MB-, 100MB-, and 1GB-dataset models specifically are available at \url{https://github.com/tylerachang/goldfish}.

\paragraph{Ambiguous or missing languages.}
Several of the FLORES and Belebele languages are either missing from Goldfish or have multiple possible Goldfish available (e.g. either the macrolanguage \texttt{que\_latn} or individual language \texttt{quy\_latn} for FLORES language \texttt{quy\_latn}).
We make the following substitutions:
\begin{itemize}[leftmargin=0.5cm,itemsep=0.0cm,topsep=0.1cm]
\item \texttt{taq\_tfng} $\to$ \texttt{None}, \newline
\texttt{tzm\_tfng} $\to$ \texttt{None}. \newline
None of the language models evaluated are trained on these languages, and no Goldfish are trained with the Tifinagh (\texttt{tfng}) script.
\item 
\texttt{awa\_deva} $\to$ \texttt{hin\_deva}, \newline
\texttt{kam\_latn} $\to$ \texttt{kik\_latn}. \newline
\texttt{kas\_arab} $\to$ \texttt{urd\_arab}, \newline
\texttt{mni\_beng} $\to$ \texttt{ben\_beng}, \newline
\texttt{nus\_latn} $\to$ \texttt{din\_latn}, \newline
\texttt{taq\_latn} $\to$ \texttt{kab\_latn}. \newline
Here, we use the closest relative in Goldfish that uses the same script.
\item 
\texttt{ace\_arab} $\to$ \texttt{urd\_arab}, \newline
\texttt{arb\_latn} $\to$ \texttt{mlt\_latn}, \newline
\texttt{ben\_latn} $\to$ \texttt{hin\_latn}, \newline
\texttt{bjn\_arab} $\to$ \texttt{urd\_arab}, \newline
\texttt{min\_arab} $\to$ \texttt{urd\_arab}, \newline
\texttt{npi\_latn} $\to$ \texttt{hin\_latn}, \newline
\texttt{sin\_latn} $\to$ \texttt{hin\_latn}, \newline
\texttt{urd\_latn} $\to$ \texttt{hin\_latn}. \newline
These are languages that are missing from Goldfish and that are written in a nonstandard script for the language (e.g. Arabic in Latin script). We use the closest relative in Goldfish that uses that script.
\item
\texttt{acm\_arab} $\to$ \texttt{arb\_arab}, \newline
\texttt{acq\_arab} $\to$ \texttt{arb\_arab}, \newline
\texttt{aeb\_arab} $\to$ \texttt{arb\_arab}, \newline
\texttt{ajp\_arab} $\to$ \texttt{arb\_arab}, \newline
\texttt{als\_latn} $\to$ \texttt{sqi\_latn}, \newline
\texttt{ars\_arab} $\to$ \texttt{arb\_arab}, \newline
\texttt{ary\_arab} $\to$ \texttt{arb\_arab}, \newline
\texttt{ayr\_latn} $\to$ \texttt{aym\_latn}, \newline
\texttt{azb\_arab} $\to$ \texttt{aze\_arab}, \newline
\texttt{azj\_latn} $\to$ \texttt{aze\_latn}, \newline
\texttt{dik\_latn} $\to$ \texttt{din\_latn}, \newline
\texttt{gaz\_latn} $\to$ \texttt{orm\_latn}, \newline
\texttt{khk\_cyrl} $\to$ \texttt{mon\_cyrl}, \newline
\texttt{kmr\_latn} $\to$ \texttt{kur\_latn}, \newline
\texttt{lvs\_latn} $\to$ \texttt{lav\_latn}, \newline
\texttt{npi\_deva} $\to$ \texttt{nep\_deva}, \newline
\texttt{ory\_orya} $\to$ \texttt{ori\_orya}, \newline
\texttt{pbt\_arab} $\to$ \texttt{pus\_arab}, \newline
\texttt{plt\_latn} $\to$ \texttt{mlg\_latn}, \newline
\texttt{quy\_latn} $\to$ \texttt{que\_latn}, \newline
\texttt{swh\_latn} $\to$ \texttt{swa\_latn}, \newline
\texttt{uzn\_latn} $\to$ \texttt{uzb\_latn}, \newline
\texttt{ydd\_hebr} $\to$ \texttt{yid\_hebr}, \newline
\texttt{yue\_hant} $\to$ \texttt{zho\_hant}, \newline
\texttt{zsm\_latn} $\to$ \texttt{msa\_latn}. \newline
These languages map to multiple different Goldfish languages or are individual languages within a macrolanguage code included in Goldfish. When the option is available, we use the Goldfish language with more data.
\end{itemize}

\onecolumn
\footnotesize

\onecolumn
\definecolor{oscar}{HTML}{FF7F0E}
\definecolor{nllb}{HTML}{1F77B4}
\definecolor{madlad400}{HTML}{2CA02C}
\definecolor{glot500}{HTML}{D62728}
\definecolor{other}{HTML}{7F7F7F}
\footnotesize
\begin{center}
\setlength\tabcolsep{0.1cm}
\begin{longtable}[width=0.9\textwidth]{|lrrrrrl|}
\multicolumn{7}{l}{\textbf{Table 5:} Goldfish languages with corresponding dataset sizes. Horizontal lines separate languages with at least} \\
\multicolumn{7}{l}{1000MB, 100MB, 10MB, and 5MB of available data.} \\
\multicolumn{7}{l}{} \\
\hline
\vspace{-0.25cm} & & & & & & \\
\endfirsthead
\hline
\vspace{-0.25cm} & & & & & & \\
\endhead
\hline
\endfoot
\label{tab:lang-list}
\textbf{Language} \phantom{\rule{0.2cm}{6pt}} & \textbf{Language} & \textbf{Script} & \textbf{Byte} & \textbf{Scaled} & \textbf{Tokens} & \textbf{Dataset Proportions} \\
& \textbf{(ISO 639-3)} & \textbf{(ISO 15924)} & \textbf{Premium} & \textbf{MB} & & \\
& & & & & & \hspace{0.2cm}{\color{oscar}\rule{5pt}{5pt}} OSCAR \\
& & & & & & \hspace{0.2cm}{\color{nllb}\rule{5pt}{5pt}} NLLB \\
& & & & & & \hspace{0.2cm}{\color{madlad400}\rule{5pt}{5pt}} MADLAD-400 \\
& & & & & & \hspace{0.2cm}{\color{glot500}\rule{5pt}{5pt}} Glot500 \\
& & & & & & \hspace{0.2cm}{\color{other}\rule{5pt}{5pt}} Other \\
\vspace{-0.25cm} & & & & & & \\
\end{longtable}
\end{center}

\end{document}